\documentclass[11pt,a4paper]{article}
\usepackage[hyperref]{acl2018}
\usepackage{times}
\usepackage{latexsym}

\usepackage{url}
\usepackage{times}
\usepackage[normalem]{ulem}

\usepackage[utf8]{inputenc}
\usepackage[small]{caption}
\usepackage{graphicx}
\usepackage{booktabs}
\usepackage{mathrsfs}
\usepackage{amsmath}
\usepackage{hyperref}
\usepackage{amssymb}
\usepackage{balance}

\usepackage{xcolor}

\aclfinalcopy 

\title{Dating Documents using Graph Convolution Networks}

 \author{Shikhar Vashishth \\
 	IISc Bangalore \\
 	{\small {\tt shikhar@iisc.ac.in}} \And
 	Shib Sankar Dasgupta \\
 	IISc Bangalore \\
 	{\small {\tt shibd@iisc.ac.in}} \And
 	Swayambhu Nath Ray \\
 	IISc Bangalore \\
 	{\small {\tt swayambhuray@iisc.ac.in}} \And
 	Partha Talukdar \\
 	IISc Bangalore \\
 	{\small {\tt ppt@iisc.ac.in}}
 }
 
\date{}

\newcommand{\refeqn}[1]{Equation \ref{#1}}
\newcommand{\reffig}[1]{Figure \ref{#1}}
\newcommand{\reftbl}[1]{Table \ref{#1}}
\newcommand{\refsec}[1]{Section \ref{#1}}

\newcommand{\m}[1]{\mathcal{#1}}
\newcommand{\method}{NeuralDater}

\newcommand*{\Scale}[2][4]{\scalebox{#1}{$#2$}}%
\begin{document}
\maketitle
\begin{abstract}
	Document date is essential for many important tasks, such as document retrieval, summarization, event detection, etc. While existing approaches for these tasks assume accurate knowledge of the document date, this is not always available, especially for arbitrary documents from the Web. Document Dating is a challenging problem which requires inference over the temporal structure of the document. Prior document dating systems have largely relied on handcrafted features while ignoring such document-internal structures. In this paper, we propose \method{}, a Graph Convolutional Network (GCN) based document dating approach which jointly exploits syntactic and temporal graph structures of document in a principled way. To the best of our knowledge, this is the first application of deep learning for the problem of document dating. Through extensive experiments on real-world datasets, we find that \method{} significantly outperforms state-of-the-art baseline by 19\% absolute (45\% relative) accuracy points.
\end{abstract}

\section{Introduction}
\label{sec:introduction}

Date of a document, also referred to as the Document Creation Time (DCT), is at the core of many important tasks, such as, information retrieval \cite{ir_time_usenix,ir_time_li,ir_time_dakka}, temporal reasoning \cite{temp_reasoner1,temp_reasoner2}, text summarization \cite{text_summ_time}, event detection \cite{event_detection}, and analysis of historical text \cite{history_time}, among others. In all such tasks, the document date is assumed to be available and also accurate -- a strong assumption, especially for arbitrary documents from the Web. Thus, there is a need to automatically predict the date of a document based on its content. This problem is referred to as \emph{Document Dating}.

\begin{figure}[t]
	\begin{minipage}{1in}
		\includegraphics[scale=0.5]{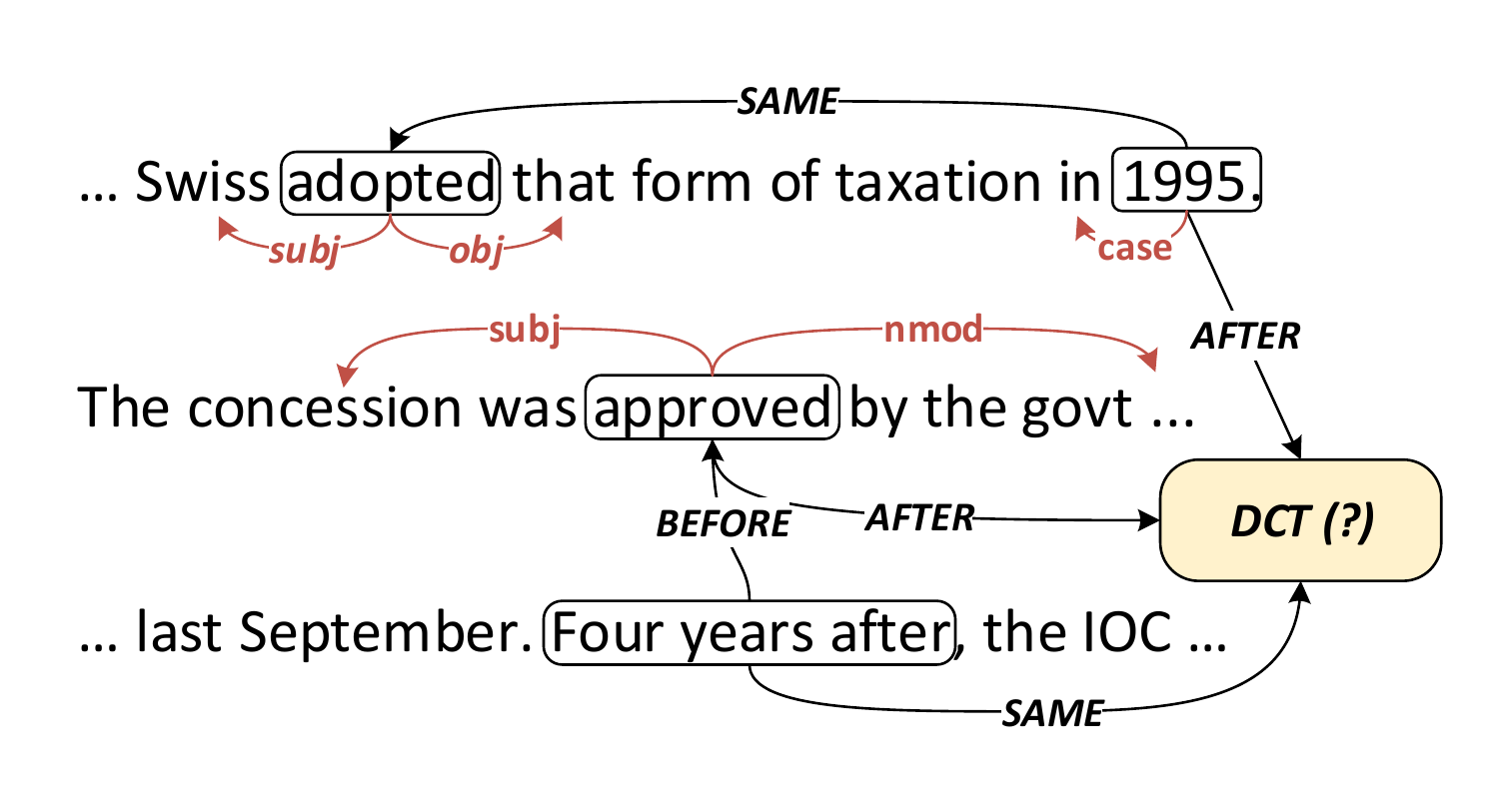}
	\end{minipage} \\
	\begin{minipage}{1in}
		\includegraphics[width=3in]{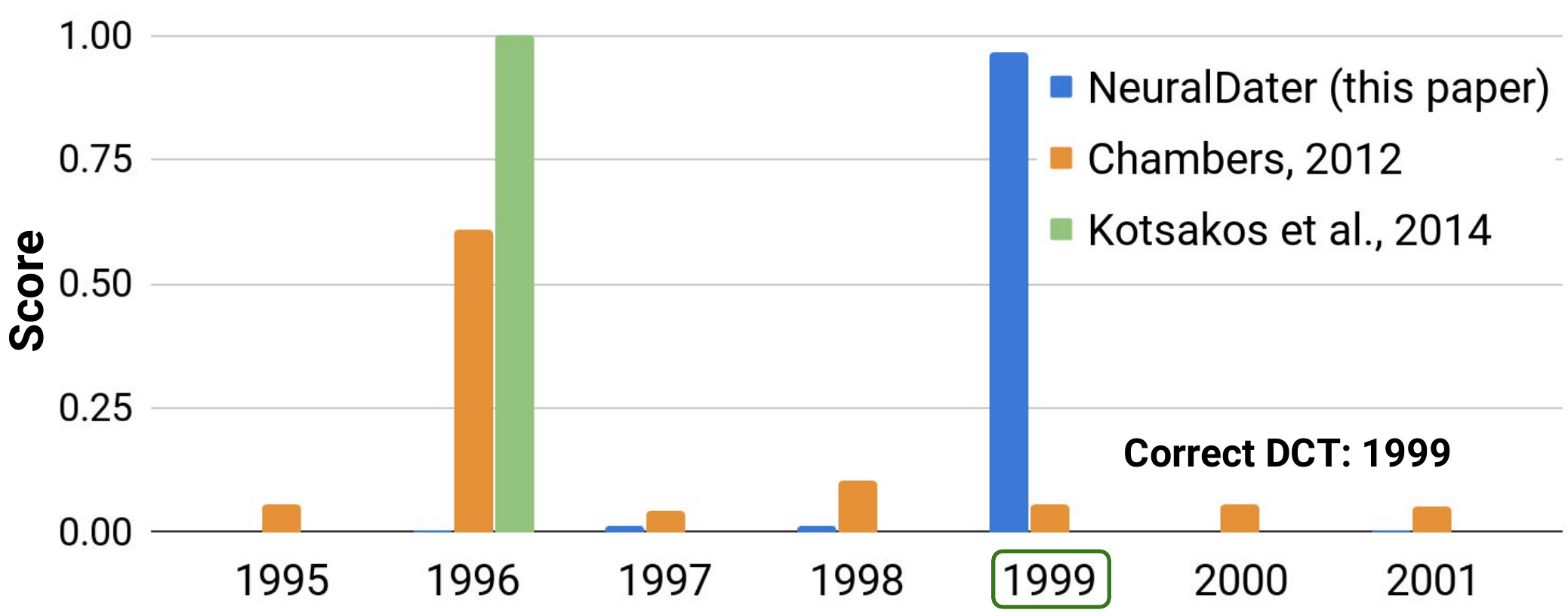}
	\end{minipage}
	\caption{\label{fig:motivation}\textbf{Top:} An example document annotated with syntactic and temporal dependencies. In order to predict the right value of 1999 for the Document Creation Time (DCT), inference over these document structures is necessary. \textbf{Bottom:} Document date prediction by two state-of-the-art-baselines and \method{}, the method proposed in this paper. While the two previous methods are getting misled by the temporal expression (\textit{1995}) in the document, \method{} is able to use the syntactic and temporal structure of the document to predict the right value (\textit{1999}). 
	}
\end{figure}

\begin{figure*}[!t]
	\centering
	\fbox{\includegraphics[width=6in]{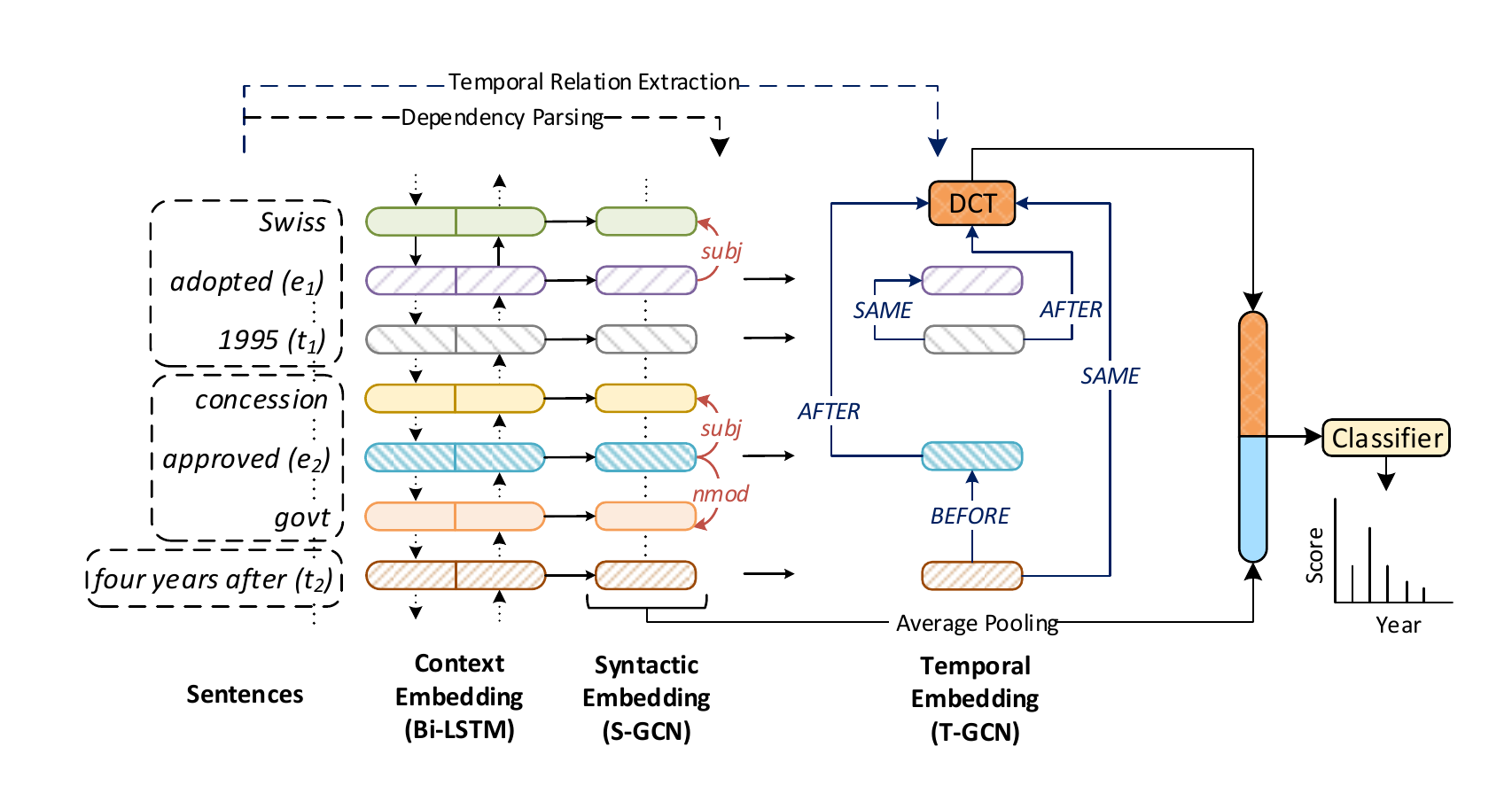}}
	\caption{\label{fig:overview}Overview of \method{}. \method{} exploits syntactic and temporal structure in a document to learn effective representation, which in turn are used to predict the  document time. \method{} uses a Bi-directional LSTM (Bi-LSTM), two Graph Convolution Networks (GCN) -- one over the dependency tree and the other over the document's temporal graph -- along with a softmax classifier, all trained end-to-end jointly. Please see \refsec{sec:method} for more details.}	
\end{figure*}

Initial attempts on automatic document dating started with generative models by \cite{de_jong05}. This model is
later improved by \cite{Kanhabua:2008:ITL:1429852.1429902} who incorporate additional features such as POS tags, collocations, etc. \citet{Chambers:2012:LDT:2390524.2390539} shows significant improvement over these prior efforts through their discriminative models using handcrafted temporal features. \citet{Kotsakos:2014:BAD:2600428.2609495} propose a statistical approach for document dating exploiting term burstiness \cite{Lappas:2009:BSD:1557019.1557075}. 

Document dating is a challenging problem which requires extensive reasoning over the temporal structure of the document. Let us motivate this through an example shown in \reffig{fig:motivation}. In the document, \textit{four years after} plays a crucial role in identifying the creation time of the document. The existing approaches give higher confidence for timestamp immediate to the year mention \textit{1995}. \method{} exploits the syntactic and temporal structure of the document to predict the right timestamp (1999) for the document. With the exception of \cite{Chambers:2012:LDT:2390524.2390539}, all prior works on the document dating problem ignore such informative temporal structure within the document.

Research in document event extraction and ordering have made it possible to extract such temporal structures involving events, temporal expressions, and the (unknown) document date in a document \cite{catena_paper,Chambers14}. While methods to perform reasoning over such structures exist \cite{tempeval07,tempeval10,tempeval13,tempeval15,timebank03}, none of them have exploited advances in deep learning \cite{alexnet,microsoft_speech,deep_learning_book}. In particular, recently proposed Graph Convolution Networks (GCN) \cite{Defferrard:2016:CNN:3157382.3157527,kipf2016semi} have emerged as a way to learn graph representation while encoding structural information and constraints represented by the graph. We adapt GCNs for the document dating problem and make the following contributions:


\begin{itemize}
	\item We propose \method{}, a Graph Convolution Network (GCN)-based approach for document dating. To the best of our knowledge, this is the first application of GCNs, and more broadly deep neural network-based methods, for the document dating problem.
	\item NeuralDater is the first document dating approach which exploits syntactic as well temporal structure of the document, all within a principled joint model.
	\item Through extensive experiments on multiple real-world datasets, we demonstrate \method{}'s effectiveness over state-of-the-art baselines.
\end{itemize}

\method{}'s source code and datasets used in the paper are available at \url{http://github.com/malllabiisc/NeuralDater}.
\section{Related Work}
\label{sec:related_work}

{\bf Automatic Document Dating}:
\citet{de_jong05} propose the first approach for automating document dating through a statistical language model. \citet{Kanhabua:2008:ITL:1429852.1429902} further extend this work by incorporating semantic-based preprocessing and temporal entropy \cite{temporal_entropy} based term-weighting. 
\citet{Chambers:2012:LDT:2390524.2390539} proposes a MaxEnt based discriminative model trained on hand-crafted temporal features. He also proposes a model to learn probabilistic constraints between year mentions and the actual creation time of the document. We draw inspiration from his work for exploiting temporal reasoning for document dating. 
\citet{Kotsakos:2014:BAD:2600428.2609495} propose a purely statistical method which considers lexical similarity alongside burstiness \cite{Lappas:2009:BSD:1557019.1557075} of terms for dating documents. To the best of our knowledge, \method{}, our proposed method,  is the first method to utilize deep learning techniques for the document dating problem.



{\bf Event Ordering Systems}: Temporal ordering of events is a vast research topic in NLP. The problem is posed as a temporal relation classification between two given temporal entities. Machine Learned classifiers and well crafted linguistic features for this task are used in \cite{Chambers:2007:CTR:1557769.1557820, E14-1033}. \citet{N13-1112} use a hybrid approach by adding 437 hand-crafted rules. \citet{Chambers:2008:JCI:1613715.1613803, P09-1046} try to classify with many more temporal constraints, while utilizing integer linear programming and Markov logic. 

CAEVO, a CAscading EVent Ordering architecture \cite{Chambers14} use sieve-based architecture 
\cite{sieve_architecture} 
for temporal event ordering for the first time. They mix multiple learners according to their precision based ranks and use transitive closure for maintaining consistency of temporal graph. \citet{catena_paper} recently propose CATENA (CAusal and TEmporal relation extraction from NAtural language texts), the first integrated system for the temporal and causal relations extraction between pre-annotated events and time expressions. They also incorporate sieve-based architecture which outperforms existing methods in temporal relation classification domain. We make use of CATENA for temporal graph construction in our work. 

{\bf Graph Convolutional Networks (GCN)}:
GCNs generalize Convolutional Neural Network (CNN) over graphs. GCN is introduced by \cite{gcn_first_work}, and later extended by \cite{Defferrard:2016:CNN:3157382.3157527} with efficient localized filter approximation in spectral domain. \citet{kipf2016semi} propose a first-order approximation of localized filters through layer-wise propagation rule. GCNs over syntactic dependency trees have been recently exploited in the field of semantic-role labeling \cite{gcn_srl}, neural machine translation \cite{gcn_nmt}, event detection \cite{gcn_event}, relation extraction \cite{reside2018}. In our work, we successfully use GCNs for document dating.

\section{Background: Graph Convolution Networks (GCN)}
\label{sec:background}


In this section, we provide an overview of Graph Convolution Networks (GCN) \cite{kipf2016semi}. GCN learns an embedding for each node of the graph it is applied over. We first present GCN for undirected graphs and then move onto GCN for directed graph setting. 

\subsection{GCN on Undirected Graph}
\label{sec:undirected_gcn}

Let $\m{G} = (\m{V}, \m{E})$ be an undirected graph, where $\m{V}$ is a set of $n$ vertices and $\m{E}$ the set of edges. The input feature matrix  $\m{X} \in \mathbb{R}^{n \times m}$ whose rows are input representation of node $u$, $x_{u} \in \mathbb{R}^{m}\text{, }\forall u \in \m{V}$. The output hidden representation $h_v \in \mathbb{R}^{d}$ of a node $v$ after a single layer of graph convolution operation can be obtained by considering only the immediate neighbors of  $v$. This can be formulated as:
$$h_{v} = f\left(\sum_{u \in \m{N}(v)}\left(W x_{u} + b\right)\right),~~~\forall v \in \m{V} .$$
Here, model parameters $W \in \mathbb{R}^{d \times m}$ and $b \in \mathbb{R}^{d}$ are learned in a task-specific setting using first-order gradient optimization. $\m{N}(v)$ refers to the set of neighbors of $v$ and $f$ is any non-linear activation function. We have used ReLU as the activation function in this paper\footnote{ReLU: $f(x) = \max(0, x)$}.  

In order to capture nodes many hops away, multiple GCN layers may be stacked one on top of another. In particular, $h^{k+1}_{v}$, representation of node $v$  after $k^{th}$ GCN layer can be formulated as:
$$h^{k+1}_{v} = f\left(\sum_{u \in \m{N}(v)}\left(W^{k} h^{k}_{u} + b^{k}\right)\right), \forall v \in \m{V} . $$
where $h^{k}_{u}$ is the input to the $k^{th}$ layer.

\subsection{GCN on Labeled and Directed Graph}
\label{sec:directed_gcn}

In this section, we consider GCN formulation over graphs where each edge is labeled as well as directed. In this setting, an edge from node $u$ to $v$ with label $l(u, v)$ is denoted as $(u, v, l(u, v))$. While a few recent works focus on GCN over directed graphs \cite{gcn_summ,gcn_srl}, none of them consider labeled edges. We handle both direction and label by incorporating label and direction specific filters.

Based on the assumption that the information in a directed edge need not only propagate along its direction, following \cite{gcn_srl} we define an updated edge set $\m{E'}$ which expands the original set $\m{E}$ by incorporating inverse, as well self-loop edges.
\begin{multline}	
\m{E'} = \m{E} \cup \{(v,u,l(u,v)^{-1})~|~ (u,v,l(u,v)) \in \m{E}\}  \\
\cup \{(u, u, \top)~|~u \in \m{V})\} \label{eqn:updated_edges}.
\end{multline}

Here, $l(u,v)^{-1}$ is the inverse edge label corresponding to label $l(u,v)$, and $\top$ is a special empty relation symbol for self-loop edges. We now define $h_{v}^{k+1}$ as the embedding of node $v$ after $k^{th}$ GCN layer applied over the directed and labeled graph as:
\begin{equation}
h_{v}^{k+1} = f \left(\sum_{u \in \m{N}(v)}\left(W^{k}_{l(u,v)}h_{u}^{k} + b^{k}_{l(u,v)}\right)\right).
\label{eqn:gcn_diretected}
\end{equation}

We note that the parameters $W^{k}_{l(u,v)}$ and $b^{k}_{l(u,v)}$ in this case are edge label specific.

\subsection{Incorporating Edge Importance}
\label{sec:gating}

In many practical settings, we may not want to give equal importance to all the edges. For example, in case of automatically constructed graphs, some of the edges may be erroneous and we may want to automatically learn to discard them. Edge-wise gating may be used in a GCN to give importance to relevant edges and subdue the noisy ones. \citet{gcn_event, gcn_srl} used gating for similar reasons and obtained high performance gain. At $k^{th}$ layer, we compute gating value for a particular edge $(u,v, l(u,v))$ as:
\[
g^{k}_{u,v} = \sigma \left( h^{k}_u \cdot \hat{w}^{k}_{l(u,v)} + \hat{b}^{k}_{l(u,v)} \right),
\]
where, $\sigma(\cdot)$ is the sigmoid function, $\hat{w}^{k}_{l(u,v)}$ and $ \hat{b}^{k}_{l(u,v)}$ are label specific gating parameters. Thus, gating helps to make the model robust to the noisy labels and directions of the input graphs. GCN embedding of a node while incorporating edge gating may be computed as follows.
\[
h^{k+1}_{v} = f\left(\sum_{u \in \m{N}(v)} g^{k}_{u,v} \times \left({W}^{k}_{l(u,v)} h^{k}_{u} + b^{k}_{l(u,v)}\right)\right).
\]

\section{\method{} Overview}
\label{sec:method}

The Documents Dating problem may be cast as a multi-class classification problem \cite{Kotsakos:2014:BAD:2600428.2609495,Chambers:2012:LDT:2390524.2390539}. In this section, we present an overview of \method{}, the document dating system proposed in this paper. Architectural overview of \method{} is shown in \reffig{fig:overview}.

\method{} is a deep learning-based multi-class classification system. It takes in a document as input and returns its predicted date as output by exploiting the syntactic and temporal structure of document. 

\method{} network consists of three layers which learn an  embedding for the Document Creation Time (DCT) node corresponding to the document. This embedding is then fed to a softmax classifier which produces a distribution over timestamps. Following prior research \cite{Chambers:2012:LDT:2390524.2390539,Kotsakos:2014:BAD:2600428.2609495}, we work with year granularity for the experiments in this paper.  We, however, note that NeuralDater can be trained for finer granularity with appropriate training data. The \method{} network is trained end-to-end using training data. We briefly present \method{}'s various components below. Each component is described in greater detail in subsequent sections.

\begin{itemize}
	\item \textbf{Context Embedding}: In this layer, \method{} uses a Bi-directional LSTM (Bi-LSTM) to learn embedding for each token in the document. Bi-LSTMs have been shown to be quite effective in capturing local context inside token embeddings \cite{Sutskever:2014:SSL:2969033.2969173}.
	\item \textbf{Syntactic Embedding}: In this step, \method{} revises token embeddings from the previous step by running a GCN over the dependency parses of sentences in the document. We refer to this GCN as \textbf{Syntactic GCN} or \textbf{S-GCN}. While the Bi-LSTM captures immediate local context in token embeddings, S-GCN augments them by capturing syntactic context. 
	\item \textbf{Temporal Embedding}: In this step, \method{} further refines embeddings learned by S-GCN to incorporate cues from temporal structure of event and times in the document. \method{} uses state-of-the-art causal and temporal relation extraction algorithm \cite{catena_paper} for extracting temporal graph for each document. A GCN is then run over this temporal graph to refine the embeddings from the previous layer. We refer to this GCN as \textbf{Temporal GCN} or \textbf{T-GCN}. In this step, a special DCT node is introduced whose embedding is also learned by the T-GCN.
	\item \textbf{Classifier}: Embedding of the DCT node along with average pooled embeddings learned by S-GCN are fed to a fully connected softmax classifier which makes the final prediction about the date of the document.
\end{itemize}

Even though the previous discussion is presented in a sequential manner, the whole network is trained in a joint end-to-end manner using backpropagation.

%
%
%
\section{\method{} Details}

In this section, we present detailed description of various components of \method{}.

\subsection{Context Embedding (Bi-LSTM)}
\label{sec:et_gcn}

Let us consider a document $D$ with $n$ tokens  $w_1, w_2, ..., w_n$.
We first represent each token by a $k$-dimensional word embedding. For the experiments in this paper, we use GloVe \cite{glove} embeddings. These token embeddings are stacked together to get the document representation $\m{X} \in \mathbb{R}^{n \times k}$. We then employ a Bi-directional LSTM (Bi-LSTM) \cite{lstm_1997} on the input matrix $\m{X}$ to obtain contextual embedding for each token. After stacking contextual embedding of all these tokens, we get the new document representation matrix $\m{H}^{cntx} \in \mathbb{R}^{n \times r_{cntx}}$. In this new representation, each token is represented in a $r_{cntx}$-dimensional space. Our choice of LSTMs for learning contextual embeddings for tokens is motivated by the previous success of LSTMs in this task \cite{Sutskever:2014:SSL:2969033.2969173}.

\subsection{Syntactic Embedding (S-GCN)} 
\label{sec:syntax_gcn}

While the Bi-LSTM is effective at capturing immediate local context of a token, it may not be as effective in capturing longer range dependencies among words in a sentence. For example, in \reffig{fig:motivation}, we would like the embedding of token \textit{approved} to be directly affected by \textit{govt}, even though they are not immediate neighbors. A dependency parse may be used to capture such longer-range connections. In fact, similar features were exploited by \cite{Chambers:2012:LDT:2390524.2390539} for the document dating problem. \method{} captures such longer-range information by using another GCN run over the syntactic structure of the document. We describe this in detail below.

The context embedding, $\m{H}^{cntx} \in \mathbb{R}^{n \times r_{cntx}}$ learned in the previous step is used as input to this layer. For a given document, we first extract its syntactic dependency structure by applying the Stanford CoreNLP's dependency parser \cite{stanford_corenlp} on each sentence in the document individually. We now employ the Graph Convolution Network (GCN) over this dependency graph using the GCN formulation presented in \refsec{sec:directed_gcn}. We call this GCN the Syntactic GCN or S-GCN, as mentioned in \refsec{sec:method}.

Since S-GCN operates over the dependency graph and uses \refeqn{eqn:gcn_diretected} for updating embeddings, the number of parameters in S-GCN is directly proportional to the number of dependency edge types. Stanford CoreNLP's dependency parser returns 55 different dependency edge types. This large number of edge types is going to significantly over-parameterize S-GCN, thereby increasing the possibility of overfitting. In order to address this, we use only three edge types in S-GCN. For each edge connecting nodes $w_i$ and $w_j$ in $\m{E'}$ (see \refeqn{eqn:updated_edges}), we determine its new type $L(w_i, w_j)$ as follows:
\begin{itemize}
	\item $L(w_i, w_j) = \rightarrow$ if $(w_i, w_j, l(w_i, w_j)) \in \m{E'}$, i.e., if the edge is an original dependency parse edge
	\item $L(w_i, w_j) = \leftarrow$ if $(w_i, w_j, l(w_i, w_j)^{-1}) \in \m{E'}$, i.e., if the edges is an inverse edge	
	\item $L(w_i, w_j) = \top$ if $(w_i, w_j, \top) \in \m{E'}$, i.e., if the edge is a self-loop with $w_i = w_j$
\end{itemize}
S-GCN now estimates embedding $h^{syn}_{w_{i}} \in \mathbb{R}^{r_{syn}}$ for each token $w_{i}$ in the document using the formulation shown below.
\[
\Scale[0.88]{h^{syn}_{w_i} = f \Bigg(\sum_{w_j \in \m{N}(w_i)}\left(W_{L(w_i, w_j)}h^{cntx}_{w_j} + b_{L(w_i, w_j)}\right) \Bigg)}
\]
Please note S-GCN's use of the new edge types $L(w_i, w_j)$  above, instead of the $l(w_i, w_j)$ types used in \refeqn{eqn:gcn_diretected}. By stacking embeddings for all the tokens together, we get the new embedding matrix $\m{H}^{syn} \in \mathbb{R}^{n \times r_{syn}}$ representing the document.

\textbf{AveragePooling}: We obtain an embedding $h_{D}^{avg}$ for the whole document by average pooling of every token representation.
\begin{equation}
h_{D}^{avg} = \frac{1}{n} \sum_{i = 1}^{n} h_{w_i}^{syn}
\label{eqn:avg-pool}.
\end{equation}


\subsection{Temporal Embedding (T-GCN)}
\label{sec:t-gcn}

In this layer, \method{} exploits temporal structure of the document to learn an embedding for the Document Creation Time (DCT) node of the document. First, we describe the construction of temporal graph, followed by GCN-based embedding learning over this graph.

\textbf{Temporal Graph Construction}: \method{} uses Stanford's SUTime tagger \cite{sutime_paper} for date normalization and the event extraction classifier of \cite{Chambers14} for event detection. The annotated document is then passed to CATENA \cite{catena_paper}, current state-of-the-art temporal and causal relation extraction algorithm, to obtain a temporal graph for each document. 
Since our task is to predict the creation time of a given document, we supply DCT as unknown to CATENA. We hypothesize that the temporal relations extracted in absence of DCT are helpful for document dating and we indeed find this to be true, as shown in Section \ref{sec:results}. Temporal graph is a directed graph, where nodes correspond to events, time mentions, and the Document Creation Time (DCT). Edges in this graph represent causal and temporal relationships between them. Each edge is attributed with a label representing the type of the temporal relation. CATENA outputs 9 different types of temporal relations, out of which we selected five types, viz.,  \textit{AFTER}, \textit{BEFORE}, \textit{SAME}, \textit{INCLUDES}, and  \textit{IS\_INCLUDED}. The remaining four types were ignored as they were substantially infrequent. 

Please note that the temporal graph may involve only a small number of tokens in the document. For example, in the temporal graph in \reffig{fig:overview}, there are a total of 5 nodes: two temporal expression nodes (\textit{1995} and \textit{four years after}), two event nodes (\textit{adopted} and \textit{approved}), and a special DCT node. This graph also consists of temporal relation edges such as (\textit{four years after}, \textit{approved}, \textit{BEFORE}). 


\textbf{Temporal Graph Convolution}: \method{} employs a GCN over the temporal graph constructed above. We refer to this GCN as the Temporal GCN or T-GCN, as mentioned in \refsec{sec:method}. T-GCN is based on the GCN formulation presented in \refsec{sec:directed_gcn}. Unlike S-GCN, here we consider label and direction specific parameters as the temporal graph consists of only five types of edges.

Let $n_T$ be the number of nodes in the temporal graph. Starting with $\m{H}^{syn}$ (\refsec{sec:syntax_gcn}), T-GCN learns a $r_{temp}$-dimensional embedding for each node in the temporal graph. Stacking all these embeddings together, we get the embedding matrix $\m{H}^{temp} \in \mathbb{R}^{n_{T} \times r_{temp}}$. T-GCN embeds the temporal constraints induced by the temporal graph in $h_{DCT}^{temp} \in \mathbb{R}^{r_{temp}}$, embedding of the DCT node of the document. 


\subsection{Classifier}
Finally, the DCT embedding $h_{DCT}^{temp}$ and average-pooled syntactic representation $h_{D}^{avg}$ (see \refeqn{eqn:avg-pool}) of document $D$ are concatenated and fed to a fully connected feed forward network followed by a softmax. This allows the \method{} to exploit context, syntactic, and temporal structure of the document to  predict the final document date $y$.
\begin{eqnarray*}
	h_{D}^{avg+temp} &=& \text{ } [h_{DCT}^{temp}~;~h_{D}^{avg}] \\ 
	p(y \vert D) &=& \mathrm{Softmax}(W \cdot h_{D}^{avg+temp} + b).
\end{eqnarray*}
\section{Experimental Setup}
\label{sec:experiments}

\begin{table}[t]
	\begin{tabular}{cccc}
		\toprule
		Datasets 	& \# Docs & Start Year & End Year\\
		\midrule
		APW 		&  675k	& 1995  & 2010 \\
		NYT			&  647k	& 1987  & 1996 \\
		\bottomrule
		\addlinespace
	\end{tabular}
	\caption{\label{tb:datasets}Details of datasets used. Please see \refsec{sec:experiments} for details.}
\end{table}


\textbf{Datasets}: We experiment on Associated Press Worldstream (APW) and New York Times (NYT) sections of Gigaword corpus \cite{gigaword5th}. The original dataset contains around 3 million documents of APW and 2 million documents of NYT from span of multiple years. From both sections, we randomly sample around 650k documents while maintaining balance among years. Documents belonging to years with substantially fewer documents are omitted. Details of the dataset can be found in Table \ref{tb:datasets}. For train, test and validation  splits, the dataset was randomly divided in 80:10:10 ratio.

\textbf{Evaluation Criteria}: Given a document, the model needs to predict the year in which the document was published. We measure performance in terms of overall accuracy of the model. 

\textbf{Baselines}: For evaluating \method{}, we compared against the following methods:

\begin{itemize}
	\item \textbf{BurstySimDater} \citet{Kotsakos:2014:BAD:2600428.2609495}:  This is a purely statistical method which uses lexical similarity and term burstiness \cite{Lappas:2009:BSD:1557019.1557075} for dating documents in arbitrary length time frame. For our experiments, we took the time frame length as 1 year. Please refer to \cite{Kotsakos:2014:BAD:2600428.2609495} for more details.
	\item \textbf{MaxEnt-Time-NER}: Maximum Entropy (MaxEnt) based classifier trained on hand-crafted temporal and Named Entity Recognizer (NER) based features. More details in \cite{Chambers:2012:LDT:2390524.2390539}. 
	\item \textbf{MaxEnt-Joint}: Refers to MaxEnt-Time-NER combined with year mention classifier as described in \cite{Chambers:2012:LDT:2390524.2390539}. 
	\item \textbf{MaxEnt-Uni-Time:} MaxEnt based discriminative model which takes bag-of-words representation of input document with normalized time expression as its features. 
	\item \textbf{CNN:} A Convolution Neural Network (CNN) \cite{cnn_paper} based text classification model proposed by \cite{yoon_kim}, which attained state-of-the-art results in several domains. 
	\item {\bf {\method{}}}: Our proposed method, refer Section \ref{sec:method}.
\end{itemize}

\textbf{Hyperparameters}: By default, edge gating (\refsec{sec:gating}) is used in all GCNs. The parameter $K$ represents the number of layers in T-GCN (\refsec{sec:t-gcn}). We use 300-dimensional GloVe embeddings and 128-dimensional hidden state for both GCNs and BiLSTM with $0.8$ dropout. We used Adam \cite{adam_optimizer} with $0.001$ learning rate for training.

\begin{table}[t]
	\centering
	\begin{tabular}{lcc}
		\toprule
		Method 			 & APW & NYT \\
		\midrule		
		\addlinespace
		BurstySimDater 		& 45.9 & 38.5 \\
		MaxEnt-Time+NER		& 52.5 & 42.3 \\
		MaxEnt-Joint		& 52.5 & 42.5 \\
		MaxEnt-Uni-Time		& 57.5 & 50.5 \\
		CNN 				& 56.3 & 50.4 \\
		\method{}			& \textbf{64.1} & \textbf{58.9} \\
		\bottomrule
	\end{tabular}
	\caption{\label{tb:result_main}Accuracies of different methods on APW and NYT datasets for the document dating problem (higher is better). \method{} significantly outperforms all other competitive baselines. This is our main result. Please see \refsec{sec:perf_comp} for more details.}

\end{table}

\begin{figure}[t]
	\centering
	\includegraphics[width=3.1in]{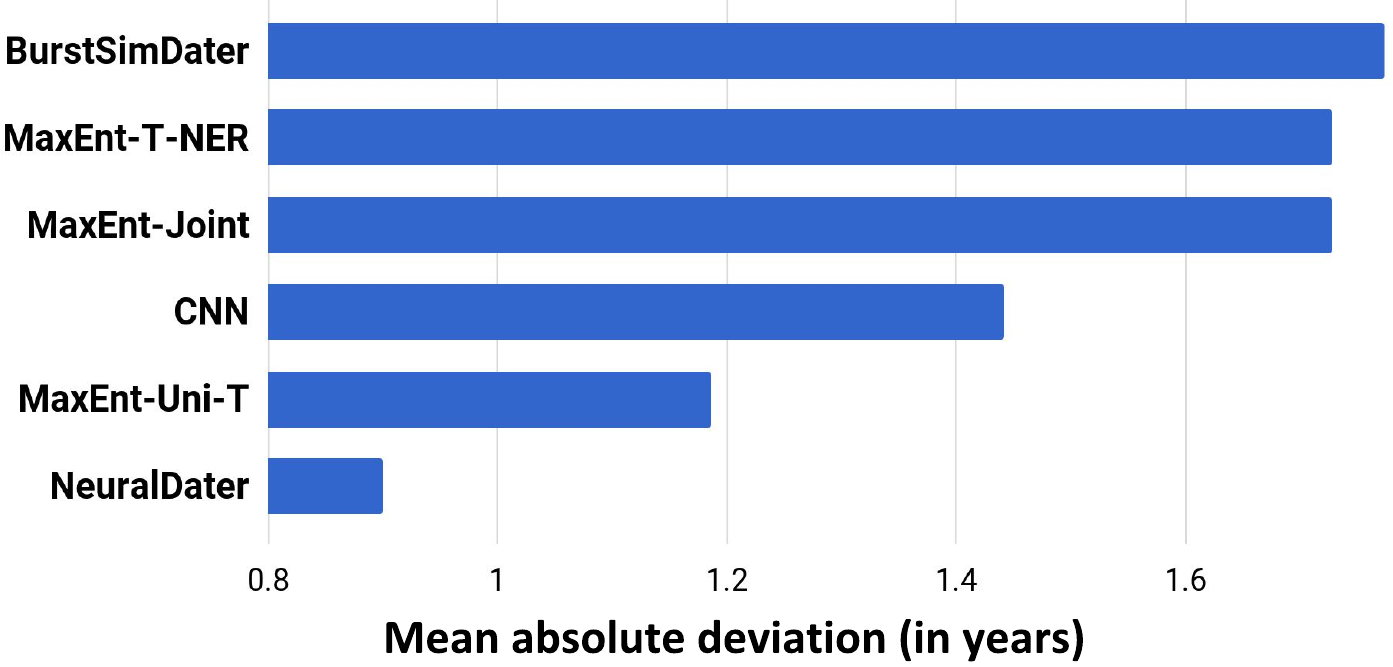}
	\caption{\label{fig:results_mean_dev}Mean absolute deviation (in years; lower is better) between a model's top prediction and the true year in the APW dataset. We find that \method{}, the proposed method, achieves the least deviation. Please see \refsec{sec:perf_comp} for details. 
	}
\end{figure}

\begin{table}[!t]
	\begin{small}
		\centering
		\begin{tabular}{lc}
			\toprule
			Method 			 & Accuracy \\
			\midrule		
			\addlinespace
			T-GCN 								& 57.3 \\
			S-GCN + T-GCN $(K=1)$				& 57.8 \\
			S-GCN + T-GCN $(K=2)$				& 58.8 \\
			S-GCN + T-GCN $(K=3)$				& \textbf{59.1} \\
			\midrule
			Bi-LSTM 							& 58.6 \\
			Bi-LSTM + CNN 						& 59.0 \\
			Bi-LSTM + T-GCN						& 60.5 \\
			Bi-LSTM + S-GCN + T-GCN (no gate)	& 62.7 \\
			Bi-LSTM + S-GCN + T-GCN $(K=1)$		& \textbf{64.1} \\
			Bi-LSTM + S-GCN + T-GCN $(K=2)$		& 63.8 \\
			Bi-LSTM + S-GCN + T-GCN $(K=3)$		& 63.3 \\
			\bottomrule
		\end{tabular}
	\caption{\label{tb:result_ablation}Accuracies of different ablated methods on the APW dataset. Overall, we observe that incorporation of context (Bi-LSTM), syntactic structure (S-GCN) and temporal structure (T-GCN) in \method{} achieves the best performance. Please see \refsec{sec:perf_comp} for details.}
\end{small}
\end{table}

\section{Results}
\label{sec:results}

\subsection{Performance Comparison}
\label{sec:perf_comp}

In order to evaluate the effectiveness of \method{}, our proposed method, we compare it against existing document dating systems and text classification models. The final results are summarized in Table \ref{tb:result_main}. Overall, we find that \method{} outperforms all other methods with a significant margin on both datasets. Compared to the previous state-of-the-art in document dating, BurstySimDater  \cite{Kotsakos:2014:BAD:2600428.2609495}, we get 19\% average absolute improvement in accuracy across both datasets. We observe only a slight gain in the performance of MaxEnt-based model (MaxEnt-Time+NER) of \cite{Chambers:2012:LDT:2390524.2390539} on combining with temporal constraint reasoner (MaxEnt-Joint). This may be attributed to the fact that the model utilizes only year mentions in the document, thus ignoring other relevant signals which might be relevant to the task. BurstySimDater performs considerably better in terms of precision compared to the other baselines,  although it significantly underperforms in accuracy. We note that NeuralDater outperforms all these prior models both in terms of precision and accuracy. We find that even generic deep-learning based text classification models, such as CNN \cite{yoon_kim}, are quite effective for the problem. However, since such a model doesn't give specific attention to temporal features in the document, its performance remains limited. From \reffig{fig:results_mean_dev}, we observe that \method{}'s top prediction achieves on average the lowest deviation from the true year.

\subsection{Ablation Comparisons}
\label{sec:ablation}

For demonstrating the efficacy of GCNs and BiLSTM for the problem, we evaluate different ablated variants of \method{} on the APW dataset. Specifically, we validate the importance of using syntactic and temporal GCNs and the effect of eliminating BiLSTM from the model. Overall results are summarized in Table \ref{tb:result_ablation}. The first block of rows in the table corresponds to the case when BiLSTM layer is excluded from \method{}, while the second block denotes the case when BiLSTM is included. We also experiment with multiple stacked layers of T-GCN (denoted by $K$) to observe its effect on the performance of the model. 

We observe that embeddings from Syntactic GCN (S-GCN) are much better than plain GloVe embeddings for T-GCN as S-GCN encodes the syntactic neighborhood information in event and time embeddings which makes them more relevant for document dating task.

Overall, we observe that including BiLSTM in the model improves performance significantly. Single BiLSTM model outperforms all the models listed in the first block of  \reftbl{tb:result_ablation}. Also, some gain in performance is observed on increasing the number of T-GCN layers ($K$) in absence of BiLSTM, although the same does not follow when BiLSTM is included in the model. This observation is consistent with \cite{gcn_srl}, as multiple GCN layers become redundant in the presence of BiLSTM. We also find that eliminating edge gating from our best model deteriorates its overall performance.

In summary, these results validate our thesis that joint incorporation of syntactic and temporal structure of a document in \method{} results in improved performance.

\subsection{Discussion and Error Analysis}
\label{sec:discussion}

\begin{figure}[t]
	\centering
	\includegraphics[width=3.1in]{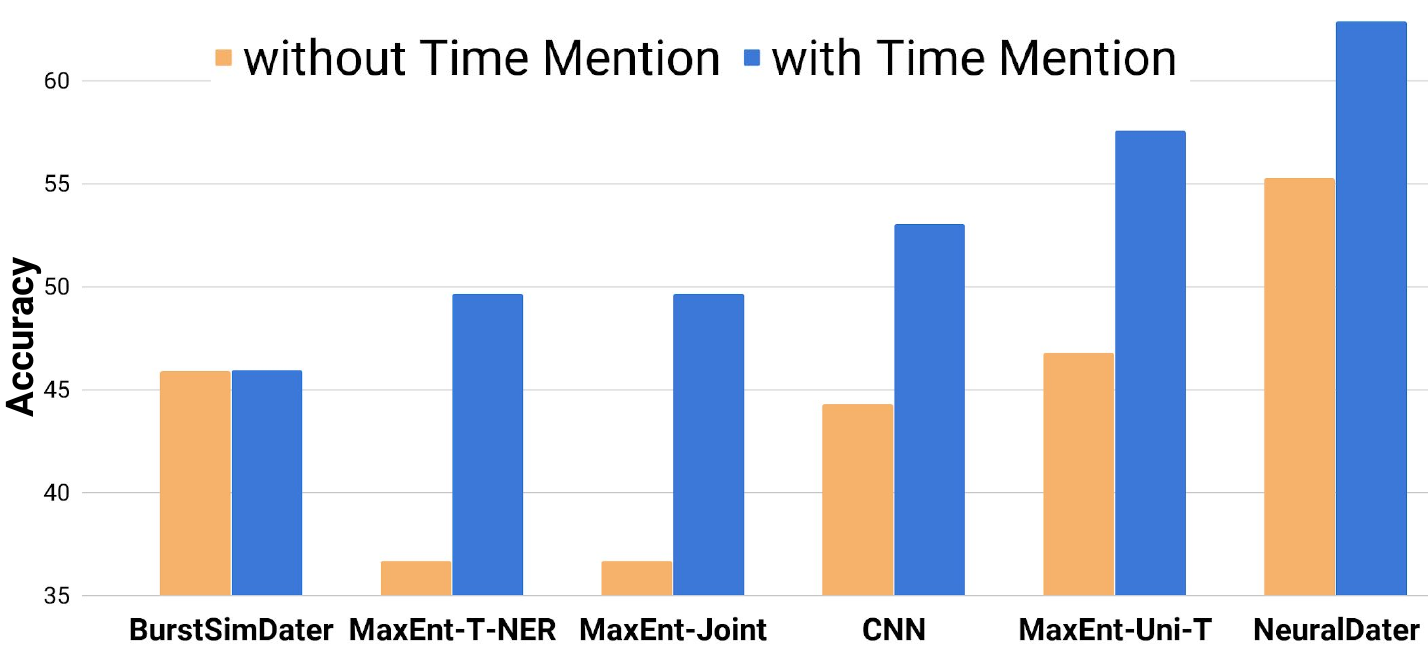}
	\caption{\label{fig:results_time_mention}Evaluating performance of different methods on dating  documents with and without time mentions. Please see \refsec{sec:discussion} for details.}
\end{figure}

In this section, we list some of our observations while trying to identify pros and cons of \method{}, our proposed method. We divided the development split of the APW dataset into two sets -- those with and without any mention of time expressions (year). We apply \method{} and other methods to these two sets of documents and report accuracies in \reffig{fig:results_time_mention}. We find that overall, \method{} performs better in comparison to the existing baselines in both scenarios. Even though the performance of \method{} degrades in the absence of time mentions, its performance is still the best relatively. 
Based on other analysis, we find that \method{} fails to identify timestamp of documents reporting local infrequent incidents without explicit time mention. \method{} becomes confused in the presence of multiple misleading time mentions; it also loses out on documents discussing events which are outside the time range of the text on which the model was trained. In future, we plan to eliminate these pitfalls by incorporating additional signals from Knowledge Graphs about entities mentioned in the document. We also plan to utilize free text temporal expression \cite{temponym_paper} in documents for improving performance on this problem. 

\section{Conclusion}
\label{sec:conclusion}

We propose \method{}, a Graph Convolutional Network (GCN) based method for document dating which exploits syntactic and temporal structures in the document in a principled way. To the best of our knowledge, this is the first application of deep learning techniques for the problem of document dating. Through extensive experiments on real-world datasets, we demonstrate the effectiveness of \method{} over existing state-of-the-art approaches. 
We are hopeful that the representation learning techniques explored in this paper will inspire further development and adoption of such techniques in the temporal information processing research community.

\section*{Acknowledgements}
We thank the anonymous reviewers for their constructive comments. This work is supported in part by the Ministry of Human Resource Development (Government of India) and by a gift from Google.

\bibliography{references}

\begin{thebibliography}{}
\expandafter\ifx\csname natexlab\endcsname\relax\def\natexlab#1{#1}\fi

\bibitem[{Allan et~al.(1998)Allan, Papka, and Lavrenko}]{event_detection}
James Allan, Ron Papka, and Victor Lavrenko. 1998.
\newblock \href{https://doi.org/10.1145/290941.290954}{On-line new event
  detection and tracking}.
\newblock In {\em Proceedings of the 21st Annual International ACM SIGIR
  Conference on Research and Development in Information Retrieval\/}. ACM, New
  York, NY, USA, SIGIR '98, pages 37--45.
\newblock
  \href{https://doi.org/10.1145/290941.290954}{https://doi.org/10.1145/290941.290954}.

\bibitem[{Bastings et~al.(2017{\natexlab{a}})Bastings, Titov, Aziz,
  Marcheggiani, and Simaan}]{gcn_nmt}
Joost Bastings, Ivan Titov, Wilker Aziz, Diego Marcheggiani, and Khalil Simaan.
  2017{\natexlab{a}}.
\newblock \href{https://www.aclweb.org/anthology/D17-1209}{Graph convolutional
  encoders for syntax-aware neural machine translation}.
\newblock In {\em Proceedings of the 2017 Conference on Empirical Methods in
  Natural Language Processing\/}. Association for Computational Linguistics,
  Copenhagen, Denmark, pages 1957--1967.
\newblock
  \href{https://www.aclweb.org/anthology/D17-1209}{https://www.aclweb.org/anthology/D17-1209}.

\bibitem[{Bastings et~al.(2017{\natexlab{b}})Bastings, Titov, Aziz,
  Marcheggiani, and Sima'an}]{gcn_event}
Joost Bastings, Ivan Titov, Wilker Aziz, Diego Marcheggiani, and Khalil
  Sima'an. 2017{\natexlab{b}}.
\newblock \href{http://arxiv.org/abs/1704.04675}{Graph convolutional encoders
  for syntax-aware neural machine translation}.
\newblock {\em CoRR\/} abs/1704.04675.
\newblock
  \href{http://arxiv.org/abs/1704.04675}{http://arxiv.org/abs/1704.04675}.

\bibitem[{Bruna et~al.(2014)Bruna, Zaremba, Szlam, and Lecun}]{gcn_first_work}
Joan Bruna, Wojciech Zaremba, Arthur Szlam, and Yann Lecun. 2014.
\newblock Spectral networks and locally connected networks on graphs.
\newblock In {\em International Conference on Learning Representations
  (ICLR2014), CBLS, April 2014\/}.

\bibitem[{Chambers(2012)}]{Chambers:2012:LDT:2390524.2390539}
Nathanael Chambers. 2012.
\newblock \href{http://dl.acm.org/citation.cfm?id=2390524.2390539}{Labeling
  documents with timestamps: Learning from their time expressions}.
\newblock In {\em Proceedings of the 50th Annual Meeting of the Association for
  Computational Linguistics: Long Papers - Volume 1\/}. Association for
  Computational Linguistics, Stroudsburg, PA, USA, ACL '12, pages 98--106.
\newblock
  \href{http://dl.acm.org/citation.cfm?id=2390524.2390539}{http://dl.acm.org/citation.cfm?id=2390524.2390539}.

\bibitem[{Chambers et~al.(2014)Chambers, Cassidy, McDowell, and
  Bethard}]{Chambers14}
Nathanael Chambers, Taylor Cassidy, Bill McDowell, and Steven Bethard. 2014.
\newblock \href{http://www.aclweb.org/anthology/Q14-1022}{Dense event ordering
  with a multi-pass architecture}.
\newblock {\em Transactions of the Association of Computational Linguistics\/}
  2:273--284.
\newblock
  \href{http://www.aclweb.org/anthology/Q14-1022}{http://www.aclweb.org/anthology/Q14-1022}.

\bibitem[{Chambers and Jurafsky(2008)}]{Chambers:2008:JCI:1613715.1613803}
Nathanael Chambers and Dan Jurafsky. 2008.
\newblock \href{http://dl.acm.org/citation.cfm?id=1613715.1613803}{Jointly
  combining implicit constraints improves temporal ordering}.
\newblock In {\em Proceedings of the Conference on Empirical Methods in Natural
  Language Processing\/}. Association for Computational Linguistics,
  Stroudsburg, PA, USA, EMNLP '08, pages 698--706.
\newblock
  \href{http://dl.acm.org/citation.cfm?id=1613715.1613803}{http://dl.acm.org/citation.cfm?id=1613715.1613803}.

\bibitem[{Chambers et~al.(2007)Chambers, Wang, and
  Jurafsky}]{Chambers:2007:CTR:1557769.1557820}
Nathanael Chambers, Shan Wang, and Dan Jurafsky. 2007.
\newblock \href{http://dl.acm.org/citation.cfm?id=1557769.1557820}{Classifying
  temporal relations between events}.
\newblock In {\em Proceedings of the 45th Annual Meeting of the ACL on
  Interactive Poster and Demonstration Sessions\/}. Association for
  Computational Linguistics, Stroudsburg, PA, USA, ACL '07, pages 173--176.
\newblock
  \href{http://dl.acm.org/citation.cfm?id=1557769.1557820}{http://dl.acm.org/citation.cfm?id=1557769.1557820}.

\bibitem[{Chang and Manning(2012)}]{sutime_paper}
Angel~X. Chang and Christopher Manning. 2012.
\newblock \href{http://www.aclweb.org/anthology/L12-1122}{Sutime: A library for
  recognizing and normalizing time expressions}.
\newblock In {\em Proceedings of the Eighth International Conference on
  Language Resources and Evaluation (LREC-2012)\/}. European Language Resources
  Association (ELRA).
\newblock
  \href{http://www.aclweb.org/anthology/L12-1122}{http://www.aclweb.org/anthology/L12-1122}.

\bibitem[{Dakka et~al.(2008)Dakka, Gravano, and Ipeirotis}]{ir_time_dakka}
Wisam Dakka, Luis Gravano, and Panagiotis~G. Ipeirotis. 2008.
\newblock \href{https://doi.org/10.1145/1458082.1458320}{Answering general time
  sensitive queries}.
\newblock In {\em Proceedings of the 17th ACM Conference on Information and
  Knowledge Management\/}. ACM, New York, NY, USA, CIKM '08, pages 1437--1438.
\newblock
  \href{https://doi.org/10.1145/1458082.1458320}{https://doi.org/10.1145/1458082.1458320}.

\bibitem[{{de Jong} et~al.(2005{\natexlab{a}}){de Jong}, Rode, and
  Hiemstra}]{history_time}
{Franciska M.G.} {de Jong}, H.~Rode, and Djoerd Hiemstra. 2005{\natexlab{a}}.
\newblock {\em Temporal Language Models for the Disclosure of Historical
  Text\/}, KNAW, pages 161--168.
\newblock Imported from EWI/DB PMS [db-utwente:inpr:0000003683].

\bibitem[{{de Jong} et~al.(2005{\natexlab{b}}){de Jong}, Rode, and
  Hiemstra}]{de_jong05}
{Franciska M.G.} {de Jong}, H.~Rode, and Djoerd Hiemstra. 2005{\natexlab{b}}.
\newblock {\em Temporal Language Models for the Disclosure of Historical
  Text\/}, KNAW, pages 161--168.
\newblock Imported from EWI/DB PMS [db-utwente:inpr:0000003683].

\bibitem[{Defferrard et~al.(2016)Defferrard, Bresson, and
  Vandergheynst}]{Defferrard:2016:CNN:3157382.3157527}
Micha\"{e}l Defferrard, Xavier Bresson, and Pierre Vandergheynst. 2016.
\newblock
  \href{http://dl.acm.org/citation.cfm?id=3157382.3157527}{Convolutional neural
  networks on graphs with fast localized spectral filtering}.
\newblock In {\em Proceedings of the 30th International Conference on Neural
  Information Processing Systems\/}. Curran Associates Inc., USA, NIPS'16,
  pages 3844--3852.
\newblock
  \href{http://dl.acm.org/citation.cfm?id=3157382.3157527}{http://dl.acm.org/citation.cfm?id=3157382.3157527}.

\bibitem[{D'Souza and Ng(2013)}]{N13-1112}
Jennifer D'Souza and Vincent Ng. 2013.
\newblock \href{http://www.aclweb.org/anthology/N13-1112}{Classifying temporal
  relations with rich linguistic knowledge}.
\newblock In {\em Proceedings of the 2013 Conference of the North American
  Chapter of the Association for Computational Linguistics: Human Language
  Technologies\/}. Association for Computational Linguistics, pages 918--927.
\newblock
  \href{http://www.aclweb.org/anthology/N13-1112}{http://www.aclweb.org/anthology/N13-1112}.

\bibitem[{Goodfellow et~al.(2016)Goodfellow, Bengio, and
  Courville}]{deep_learning_book}
Ian Goodfellow, Yoshua Bengio, and Aaron Courville. 2016.
\newblock {\em Deep Learning\/}.
\newblock MIT Press.
\newblock \url{http://www.deeplearningbook.org}.

\bibitem[{Hinton et~al.(2012)Hinton, Deng, Yu, Dahl, r.~Mohamed, Jaitly,
  Senior, Vanhoucke, Nguyen, Sainath, and Kingsbury}]{microsoft_speech}
G.~Hinton, L.~Deng, D.~Yu, G.~E. Dahl, A.~r.~Mohamed, N.~Jaitly, A.~Senior,
  V.~Vanhoucke, P.~Nguyen, T.~N. Sainath, and B.~Kingsbury. 2012.
\newblock \href{https://doi.org/10.1109/MSP.2012.2205597}{Deep neural networks
  for acoustic modeling in speech recognition: The shared views of four
  research groups}.
\newblock {\em IEEE Signal Processing Magazine\/} 29(6):82--97.
\newblock
  \href{https://doi.org/10.1109/MSP.2012.2205597}{https://doi.org/10.1109/MSP.2012.2205597}.

\bibitem[{Hochreiter and Schmidhuber(1997)}]{lstm_1997}
Sepp Hochreiter and J\"{u}rgen Schmidhuber. 1997.
\newblock \href{https://doi.org/10.1162/neco.1997.9.8.1735}{Long short-term
  memory}.
\newblock {\em Neural Comput.\/} 9(8):1735--1780.
\newblock
  \href{https://doi.org/10.1162/neco.1997.9.8.1735}{https://doi.org/10.1162/neco.1997.9.8.1735}.

\bibitem[{Kanhabua and
  N{\o}rv{\aa}g(2008{\natexlab{a}})}]{Kanhabua:2008:ITL:1429852.1429902}
Nattiya Kanhabua and Kjetil N{\o}rv{\aa}g. 2008{\natexlab{a}}.
\newblock Improving temporal language models for determining time of
  non-timestamped documents.
\newblock In {\em Proceedings of the 12th European Conference on Research and
  Advanced Technology for Digital Libraries\/}. Springer-Verlag, Berlin,
  Heidelberg, ECDL '08, pages 358--370.

\bibitem[{Kanhabua and N{\o}rv{\aa}g(2008{\natexlab{b}})}]{temporal_entropy}
Nattiya Kanhabua and Kjetil N{\o}rv{\aa}g. 2008{\natexlab{b}}.
\newblock Improving temporal language models for determining time of
  non-timestamped documents.
\newblock In {\em International Conference on Theory and Practice of Digital
  Libraries\/}. Springer, pages 358--370.

\bibitem[{Kim(2014)}]{yoon_kim}
Yoon Kim. 2014.
\newblock \href{https://doi.org/10.3115/v1/D14-1181}{Convolutional neural
  networks for sentence classification}.
\newblock In {\em Proceedings of the 2014 Conference on Empirical Methods in
  Natural Language Processing (EMNLP)\/}. Association for Computational
  Linguistics, pages 1746--1751.
\newblock
  \href{https://doi.org/10.3115/v1/D14-1181}{https://doi.org/10.3115/v1/D14-1181}.

\bibitem[{Kingma and Ba(2014)}]{adam_optimizer}
Diederik~P. Kingma and Jimmy Ba. 2014.
\newblock Adam: A method for stochastic optimization.
\newblock {\em CoRR\/} abs/1412.6980.

\bibitem[{Kipf and Welling(2017)}]{kipf2016semi}
Thomas~N. Kipf and Max Welling. 2017.
\newblock Semi-supervised classification with graph convolutional networks.
\newblock In {\em International Conference on Learning Representations
  (ICLR)\/}.

\bibitem[{Kotsakos et~al.(2014)Kotsakos, Lappas, Kotzias, Gunopulos, Kanhabua,
  and N{\o}rv{\aa}g}]{Kotsakos:2014:BAD:2600428.2609495}
Dimitrios Kotsakos, Theodoros Lappas, Dimitrios Kotzias, Dimitrios Gunopulos,
  Nattiya Kanhabua, and Kjetil N{\o}rv{\aa}g. 2014.
\newblock \href{https://doi.org/10.1145/2600428.2609495}{A burstiness-aware
  approach for document dating}.
\newblock In {\em Proceedings of the 37th International ACM SIGIR Conference on
  Research \&\#38; Development in Information Retrieval\/}. ACM, New York, NY,
  USA, SIGIR '14, pages 1003--1006.
\newblock
  \href{https://doi.org/10.1145/2600428.2609495}{https://doi.org/10.1145/2600428.2609495}.

\bibitem[{Krizhevsky et~al.(2012)Krizhevsky, Sutskever, and Hinton}]{alexnet}
Alex Krizhevsky, Ilya Sutskever, and Geoffrey~E. Hinton. 2012.
\newblock \href{http://dl.acm.org/citation.cfm?id=2999134.2999257}{Imagenet
  classification with deep convolutional neural networks}.
\newblock In {\em Proceedings of the 25th International Conference on Neural
  Information Processing Systems - Volume 1\/}. Curran Associates Inc., USA,
  NIPS'12, pages 1097--1105.
\newblock
  \href{http://dl.acm.org/citation.cfm?id=2999134.2999257}{http://dl.acm.org/citation.cfm?id=2999134.2999257}.

\bibitem[{Kuzey et~al.(2016)Kuzey, Setty, Str\"{o}tgen, and
  Weikum}]{temponym_paper}
Erdal Kuzey, Vinay Setty, Jannik Str\"{o}tgen, and Gerhard Weikum. 2016.
\newblock \href{https://doi.org/10.1145/2872427.2883055}{As time goes by:
  Comprehensive tagging of textual phrases with temporal scopes}.
\newblock In {\em Proceedings of the 25th International Conference on World
  Wide Web\/}. International World Wide Web Conferences Steering Committee,
  Republic and Canton of Geneva, Switzerland, WWW '16, pages 915--925.
\newblock
  \href{https://doi.org/10.1145/2872427.2883055}{https://doi.org/10.1145/2872427.2883055}.

\bibitem[{Lappas et~al.(2009)Lappas, Arai, Platakis, Kotsakos, and
  Gunopulos}]{Lappas:2009:BSD:1557019.1557075}
Theodoros Lappas, Benjamin Arai, Manolis Platakis, Dimitrios Kotsakos, and
  Dimitrios Gunopulos. 2009.
\newblock \href{https://doi.org/10.1145/1557019.1557075}{On burstiness-aware
  search for document sequences}.
\newblock In {\em Proceedings of the 15th ACM SIGKDD International Conference
  on Knowledge Discovery and Data Mining\/}. ACM, New York, NY, USA, KDD '09,
  pages 477--486.
\newblock
  \href{https://doi.org/10.1145/1557019.1557075}{https://doi.org/10.1145/1557019.1557075}.

\bibitem[{LeCun et~al.(1999)LeCun, Haffner, Bottou, and Bengio}]{cnn_paper}
Yann LeCun, Patrick Haffner, L{\'e}on Bottou, and Yoshua Bengio. 1999.
\newblock \href{http://dl.acm.org/citation.cfm?id=646469.691875}{Object
  recognition with gradient-based learning}.
\newblock In {\em Shape, Contour and Grouping in Computer Vision\/}.
  Springer-Verlag, London, UK, UK, pages 319--.
\newblock
  \href{http://dl.acm.org/citation.cfm?id=646469.691875}{http://dl.acm.org/citation.cfm?id=646469.691875}.

\bibitem[{Lee et~al.(2013)Lee, Chang, Peirsman, Chambers, Surdeanu, and
  Jurafsky}]{sieve_architecture}
Heeyoung Lee, Angel Chang, Yves Peirsman, Nathanael Chambers, Mihai Surdeanu,
  and Dan Jurafsky. 2013.
\newblock Deterministic coreference resolution based on entity-centric,
  precision-ranked rules.
\newblock {\em Comput. Linguist.\/} 39(4):885--916.

\bibitem[{Li and Croft(2003)}]{ir_time_li}
Xiaoyan Li and W.~Bruce Croft. 2003.
\newblock \href{https://doi.org/10.1145/956863.956951}{Time-based language
  models}.
\newblock In {\em Proceedings of the Twelfth International Conference on
  Information and Knowledge Management\/}. ACM, New York, NY, USA, CIKM '03,
  pages 469--475.
\newblock
  \href{https://doi.org/10.1145/956863.956951}{https://doi.org/10.1145/956863.956951}.

\bibitem[{Llid{\'o} et~al.(2001)Llid{\'o}, Berlanga, and
  Aramburu}]{temp_reasoner2}
D.~Llid{\'o}, R.~Berlanga, and M.~J. Aramburu. 2001.
\newblock Extracting temporal references to assign document event-time
  periods*.
\newblock In Heinrich~C. Mayr, Jiri Lazansky, Gerald Quirchmayr, and Pavel
  Vogel, editors, {\em Database and Expert Systems Applications\/}. Springer
  Berlin Heidelberg, Berlin, Heidelberg, pages 62--71.

\bibitem[{Llorens et~al.(2015)Llorens, Chambers, UzZaman, Mostafazadeh, Allen,
  and Pustejovsky}]{tempeval15}
Hector Llorens, Nathanael Chambers, Naushad UzZaman, Nasrin Mostafazadeh, James
  Allen, and James Pustejovsky. 2015.
\newblock Semeval-2015 task 5: Qa tempeval-evaluating temporal information
  understanding with question answering.
\newblock In {\em Proceedings of the 9th International Workshop on Semantic
  Evaluation (SemEval 2015)\/}. pages 792--800.

\bibitem[{Mani and Wilson(2000)}]{temp_reasoner1}
Inderjeet Mani and George Wilson. 2000.
\newblock \href{https://doi.org/10.3115/1075218.1075228}{Robust temporal
  processing of news}.
\newblock In {\em Proceedings of the 38th Annual Meeting on Association for
  Computational Linguistics\/}. Association for Computational Linguistics,
  Stroudsburg, PA, USA, ACL '00, pages 69--76.
\newblock
  \href{https://doi.org/10.3115/1075218.1075228}{https://doi.org/10.3115/1075218.1075228}.

\bibitem[{Manning et~al.(2014)Manning, Surdeanu, Bauer, Finkel, Bethard, and
  McClosky}]{stanford_corenlp}
Christopher~D. Manning, Mihai Surdeanu, John Bauer, Jenny Finkel, Steven~J.
  Bethard, and David McClosky. 2014.
\newblock \href{http://www.aclweb.org/anthology/P/P14/P14-5010}{The {Stanford}
  {CoreNLP} natural language processing toolkit}.
\newblock In {\em Association for Computational Linguistics (ACL) System
  Demonstrations\/}. pages 55--60.
\newblock
  \href{http://www.aclweb.org/anthology/P/P14/P14-5010}{http://www.aclweb.org/anthology/P/P14/P14-5010}.

\bibitem[{Marcheggiani and Titov(2017)}]{gcn_srl}
Diego Marcheggiani and Ivan Titov. 2017.
\newblock \href{http://arxiv.org/abs/1703.04826}{Encoding sentences with graph
  convolutional networks for semantic role labeling}.
\newblock {\em CoRR\/} abs/1703.04826.
\newblock
  \href{http://arxiv.org/abs/1703.04826}{http://arxiv.org/abs/1703.04826}.

\bibitem[{Mirza and Tonelli(2014)}]{E14-1033}
Paramita Mirza and Sara Tonelli. 2014.
\newblock \href{https://doi.org/10.3115/v1/E14-1033}{Classifying temporal
  relations with simple features}.
\newblock In {\em Proceedings of the 14th Conference of the European Chapter of
  the Association for Computational Linguistics\/}. Association for
  Computational Linguistics, pages 308--317.
\newblock
  \href{https://doi.org/10.3115/v1/E14-1033}{https://doi.org/10.3115/v1/E14-1033}.

\bibitem[{Mirza and Tonelli(2016)}]{catena_paper}
Paramita Mirza and Sara Tonelli. 2016.
\newblock \href{http://www.aclweb.org/anthology/C16-1007}{Catena: Causal and
  temporal relation extraction from natural language texts}.
\newblock In {\em Proceedings of COLING 2016, the 26th International Conference
  on Computational Linguistics: Technical Papers\/}. The COLING 2016 Organizing
  Committee, pages 64--75.
\newblock
  \href{http://www.aclweb.org/anthology/C16-1007}{http://www.aclweb.org/anthology/C16-1007}.

\bibitem[{Olson et~al.(1999)Olson, Bostic, Seltzer, and
  Berkeley}]{ir_time_usenix}
MA~Olson, K~Bostic, MI~Seltzer, and DB~Berkeley. 1999.
\newblock Usenix annual technical conference, freenix track.

\bibitem[{Parker et~al.(2011)Parker, Graff, Kong, Chen, and
  Maeda}]{gigaword5th}
Robert Parker, David Graff, Junbo Kong, Ke~Chen, and Kazuaki Maeda. 2011.
\newblock English gigaword fifth edition ldc2011t07. dvd.
\newblock {\em Philadelphia: Linguistic Data Consortium\/} .

\bibitem[{Pennington et~al.(2014)Pennington, Socher, and Manning}]{glove}
Jeffrey Pennington, Richard Socher, and Christopher~D. Manning. 2014.
\newblock \href{http://www.aclweb.org/anthology/D14-1162}{Glove: Global vectors
  for word representation}.
\newblock In {\em Empirical Methods in Natural Language Processing (EMNLP)\/}.
  pages 1532--1543.
\newblock
  \href{http://www.aclweb.org/anthology/D14-1162}{http://www.aclweb.org/anthology/D14-1162}.

\bibitem[{Pustejovsky et~al.(2003)Pustejovsky, Hanks, Sauri, See, Gaizauskas,
  Setzer, Radev, Sundheim, Day, Ferro et~al.}]{timebank03}
James Pustejovsky, Patrick Hanks, Roser Sauri, Andrew See, Robert Gaizauskas,
  Andrea Setzer, Dragomir Radev, Beth Sundheim, David Day, Lisa Ferro, et~al.
  2003.
\newblock The timebank corpus.
\newblock In {\em Corpus linguistics\/}. Lancaster, UK., volume 2003, page~40.

\bibitem[{Sutskever et~al.(2014)Sutskever, Vinyals, and
  Le}]{Sutskever:2014:SSL:2969033.2969173}
Ilya Sutskever, Oriol Vinyals, and Quoc~V. Le. 2014.
\newblock \href{http://dl.acm.org/citation.cfm?id=2969033.2969173}{Sequence to
  sequence learning with neural networks}.
\newblock In {\em Proceedings of the 27th International Conference on Neural
  Information Processing Systems - Volume 2\/}. MIT Press, Cambridge, MA, USA,
  NIPS'14, pages 3104--3112.
\newblock
  \href{http://dl.acm.org/citation.cfm?id=2969033.2969173}{http://dl.acm.org/citation.cfm?id=2969033.2969173}.

\bibitem[{UzZaman et~al.(2013)UzZaman, Llorens, Derczynski, Allen, Verhagen,
  and Pustejovsky}]{tempeval13}
Naushad UzZaman, Hector Llorens, Leon Derczynski, James Allen, Marc Verhagen,
  and James Pustejovsky. 2013.
\newblock Semeval-2013 task 1: Tempeval-3: Evaluating time expressions, events,
  and temporal relations.
\newblock In {\em Second Joint Conference on Lexical and Computational
  Semantics (* SEM), Volume 2: Proceedings of the Seventh International
  Workshop on Semantic Evaluation (SemEval 2013)\/}. volume~2, pages 1--9.

\bibitem[{Vashishth et~al.(2018)Vashishth, Joshi, Prayaga, Bhattacharyya, and
  Talukdar}]{reside2018}
Shikhar Vashishth, Rishabh Joshi, Sai~Suman Prayaga, Chiranjib Bhattacharyya,
  and Partha Talukdar. 2018.
\newblock \href{http://aclweb.org/anthology/D18-1157}{Reside: Improving
  distantly-supervised neural relation extraction using side information}.
\newblock In {\em Proceedings of the 2018 Conference on Empirical Methods in
  Natural Language Processing\/}. Association for Computational Linguistics,
  pages 1257--1266.
\newblock
  \href{http://aclweb.org/anthology/D18-1157}{http://aclweb.org/anthology/D18-1157}.

\bibitem[{Verhagen et~al.(2007)Verhagen, Gaizauskas, Schilder, Hepple, Katz,
  and Pustejovsky}]{tempeval07}
Marc Verhagen, Robert Gaizauskas, Frank Schilder, Mark Hepple, Graham Katz, and
  James Pustejovsky. 2007.
\newblock Semeval-2007 task 15: Tempeval temporal relation identification.
\newblock In {\em Proceedings of the 4th international workshop on semantic
  evaluations\/}. Association for Computational Linguistics, pages 75--80.

\bibitem[{Verhagen et~al.(2010)Verhagen, Sauri, Caselli, and
  Pustejovsky}]{tempeval10}
Marc Verhagen, Roser Sauri, Tommaso Caselli, and James Pustejovsky. 2010.
\newblock Semeval-2010 task 13: Tempeval-2.
\newblock In {\em Proceedings of the 5th international workshop on semantic
  evaluation\/}. Association for Computational Linguistics, pages 57--62.

\bibitem[{Wan(2007)}]{text_summ_time}
Xiaojun Wan. 2007.
\newblock \href{https://doi.org/10.1145/1277741.1277949}{Timedtextrank: Adding
  the temporal dimension to multi-document summarization}.
\newblock In {\em Proceedings of the 30th Annual International ACM SIGIR
  Conference on Research and Development in Information Retrieval\/}. ACM, New
  York, NY, USA, SIGIR '07, pages 867--868.
\newblock
  \href{https://doi.org/10.1145/1277741.1277949}{https://doi.org/10.1145/1277741.1277949}.

\bibitem[{Yasunaga et~al.(2017)Yasunaga, Zhang, Meelu, Pareek, Srinivasan, and
  Radev}]{gcn_summ}
Michihiro Yasunaga, Rui Zhang, Kshitijh Meelu, Ayush Pareek, Krishnan
  Srinivasan, and Dragomir~R. Radev. 2017.
\newblock Graph-based neural multi-document summarization.
\newblock In {\em Proceedings of CoNLL-2017\/}. Association for Computational
  Linguistics.

\bibitem[{Yoshikawa et~al.(2009)Yoshikawa, Riedel, Asahara, and
  Matsumoto}]{P09-1046}
Katsumasa Yoshikawa, Sebastian Riedel, Masayuki Asahara, and Yuji Matsumoto.
  2009.
\newblock \href{http://www.aclweb.org/anthology/P09-1046}{Jointly identifying
  temporal relations with markov logic}.
\newblock In {\em Proceedings of the Joint Conference of the 47th Annual
  Meeting of the ACL and the 4th International Joint Conference on Natural
  Language Processing of the AFNLP\/}. Association for Computational
  Linguistics, pages 405--413.
\newblock
  \href{http://www.aclweb.org/anthology/P09-1046}{http://www.aclweb.org/anthology/P09-1046}.

\end{thebibliography}
\balance
\bibliographystyle{acl_natbib}

\end{document}